\documentclass{article}

\usepackage[a4paper, left=3cm, right=3cm, top=2.5cm, bottom=3cm]{geometry}

\usepackage{fancyhdr}
\fancypagestyle{plain}{
\fancyhead[C]{\textit{Published in German Conference on Medical Image Computing (Bildverarbeitung für die Medizin, BVM) 2026 Proceedings}}
	
	\fancyfoot[C]{\thepage}}

\usepackage{authblk}
\usepackage{graphicx}

\usepackage{natbib}   % fixes \newblock
\usepackage{url}      % fixes \url
\usepackage[colorlinks]{hyperref}

\usepackage{float}
\usepackage{caption}

\begin{document}
\title{A Master Class on Reproducibility: A Student Hackathon on Advanced MRI Reconstruction Methods}

% --- AUTHORS & AFFILIATIONS IN ARTICLE STYLE ---

\author[1,2,3]{Lina Felsner\thanks{These authors contributed equally. Corresponding authors: lina.felsner@tum.de, s.kafali@tum.de.}}
\author[1]{Sevgi G.\ Kafali\textsuperscript{*}}
\author[1,2]{Hannah Eichhorn}
\author[1]{Agnes A.\ J.\ Leth}
\author[1]{Aidas Batvinskas}
\author[1]{Andr\'e Datchev}
\author[1]{Fabian Klemm}
\author[1]{Jan Aulich}
\author[1]{Puntika Leepagorn}
\author[1]{Ruben Klinger}
\author[1,3,4,5]{Daniel Rueckert}
\author[1,2,3,6]{Julia A.\ Schnabel}

\affil[1]{School of Computation, Information \& Technology, Technical University of Munich (TUM), Germany}
\affil[2]{Institute of Machine Learning in Biomedical Imaging, Helmholtz Munich, Germany}
\affil[3]{Munich Center for Machine Learning (MCML), Germany}
\affil[4]{School of Medicine and Health, TUM University Hospital Rechts der Isar, Germany}
\affil[5]{Department of Computing, Imperial College London, UK}
\affil[6]{School of Biomedical Engineering and Imaging Sciences, King’s College London, UK}

\date{} % no date, or \date{\today}

% --- OPTIONAL SHORT TITLE / AUTHOR LIST (manual in article) ---
% If you want a short title in the header, you can redefine \markboth:
\markboth{Felsner, Kafali et al.}{Reproducing DL MRI Reconstruction}

\maketitle

% Abstract of your paper, only for long papers
% Do NOT use \begin{abstract} ... \end{abstract} for short articles
\begin{abstract}
We report the design, protocol, and outcomes of a \emph{student reproducibility hackathon} focused on replicating the results of three influential MRI reconstruction papers: (a) MoDL, an unrolled model-based network with learned denoising~\cite{3757-Aggarwal2019MoDL}; (b) HUMUS-Net, a hybrid unrolled multi-scale CNN+Transformer architecture~\cite{3757-Fabian2022HUMUSNet}; and (c) an untrained, physics-regularized dynamic MRI method that uses a quantitative MR model for early stopping~\cite{3757-Slavkova2023Untrained}. 
We describe the setup of the hackathon and present reproduction outcomes alongside additional experiments, and we detail fundamental practices for building reproducible codebases.
\end{abstract}

\section{Introduction}
\label{3757-sec:intro}

Magnetic resonance imaging (MRI) is known for its strong soft-tissue contrast and can be used to diagnose several diseases from tumors to strokes and joint injuries. 
One of the biggest challenges in MRI is the long scan time, which is why 
accelerated MRI reconstruction has been extensively studied over the last few decades. 
This is done by using sophisticated algorithms which can mitigate acceleration artifacts.
Deep learning has transformed accelerated MRI reconstruction by casting it as a learned regularized inverse problem. 
A standard approach for high-quality reconstructions with strong physics grounding are unrolled neural networks, which explicitly interleave \emph{data-consistency} with learned \emph{priors}.
The Model-Based Deep Learning Architecture for Inverse Problems (MoDL)~\cite{3757-Aggarwal2019MoDL} introduces a supervised unrolled network that integrates a variational framework with a data-consistency term and a learned CNN to capture image redundancy.
Another unrolled network, called HUMUS-Net~\cite{3757-Fabian2022HUMUSNet} is a hybrid architecture that leverages transformers to model long-range dependencies while retaining the efficiency and inductive bias of convolutions.  
Following the growing interest in unsupervised approaches, Slavkova~\emph{et al.}~\cite{3757-Slavkova2023Untrained} proposed an untrained, i.e., self-supervised network that exploits the architectural bias and scan-specific optimization, while benefiting from physics-based regularization to reduce reliance on fully sampled ground truth. 
Even at higher acceleration rates, these previous studies all yield extraordinary image reconstruction quality, validated by extensive evaluations and experiments. 

%For computational research — including MRI reconstruction — reproducibility is a paramount concern. 
However, complex and intricate deep learning frameworks, specialized data acquisition protocols, and sophisticated training regimens, can make it challenging to verify reported results. 
Additional challenges might arise while reproducing the results due to varying software dependencies, hardware configurations, random seeds, and subtle differences in pre-processing or hyperparameter tuning. 
Therefore, best practices emphasize the use of version-controlled code, standardized computing environments, and fully executable pipelines. 
These practices are supported by pragmatic checklists and further reinforced by the Findable, Accessible, Interoperable, and Reusable (FAIR) principles~\cite{3757-wilkinson2016fair}, which extend the scope from data to also models and workflows.
Beyond FAIR, the FUTURE-AI~\cite{3757-lekadir2025future} principles offer a concise, consensus framework for trustworthy medical AI across the full lifecycle considering: Fairness, Universality, Traceability, Usability, Robustness, and Explainability. 
The guideline translates these six pillars into actionable recommendations for data curation, model development, validation, deployment, and monitoring.  
Following these principles alongside open code and defined environments could also strengthen both the credibility and reproducibility of reported results.

For computational research reproducibility has become an essential component~\cite{3757-Stupple2019Reproducibility}.
In the medical imaging community, two main conferences have been holding hackathons focusing on reproducibility: the annual meeting of International Society of Magnetic Resonance in Medicine (ISMRM)~\cite{3757-maier2021cg} and Medical Image Computing and Computer Assisted Interventions (MICCAI)~\cite{3757-balsiger2021miccai}. 
In addition, educational hackathons specifically focusing on reproducibility have emerged as a valuable tool for promoting best practices in a teaching environment. 
Initiatives such as the `Reproduced Papers' series~\cite{3757-reproducedpapers}
or classroom setting type hackathons~\cite{3757-reproduce_cvpr} highlight the growing recognition of reproducibility challenges. 
Particularly, the `reproduce CVPR Hackathon'~\cite{3757-reproduce_cvpr}, organized 
at Friedrich-Alexander-Universität Erlangen-Nürnberg (FAU) has aimed to reproduce papers from CVPR 2024. 
The FAU hackathon contained three stages: automated detection of code availability, a PhD hackathon to evaluate reproducibility effort, and a master’s course to 
%systematically 
reproduce CVPR papers.

This work documents a student hackathon organized to reproduce three selected works in the context of accelerated MRI reconstruction: HUMUS-Net, MoDL, and untrained+physics model. The hackathon has been designed to (1) help students connect theory and practice in inverse problems applied to MRI reconstruction; and (2) emphasize reproducibility and responsible use of MRI data.

\section{Methods}
\label{3757-sec:methods}

In this section we describe the setup of our reproducibility hackathon.
This hackathon has been an integral component of the `Deep Learning for Inverse Problems in Medical Imaging' master seminar at Technical University of Munich (TUM). 
Its primary objective has been to transition students from paper reading (i.e., theoretical understanding) to practical application with a hands-on experience.

The hackathon spanned a total of four hours, divided across three sessions. % (1.5 hours, 1 hour, and 1.5 hours). 
Paper selection was done by the seminar course instructors. 
During this selection, instructors ensured code and data availability including MR images and the respective raw data, as well as pre-trained model weights. 
%Seven students, originating from diverse master's degree programs, formed three teams.
Seven students from the School of Computation, Information and Technology at TUM, %CIT at TUM, \TODO{anonymize for submission.} School of Computation, Information and Technology
originating from diverse master's programs, formed three teams.
Each team selected one out of six possible papers. 
The task was to recreate one of the figures in the result section of the chosen scientific paper. 
The students were asked to utilize the provided pre-trained weights, thus bypassing the need for training models from scratch % due to time constraints.
to avoid differences in code or training set-up.
If the students could complete the task before the deadline, they were asked to show additional out-of-distribution (i.e., cross-domain) results in a test dataset for bonus credit. 
\begin{itemize}
  %A
  \item \emph{Team A:} The first team consisted of two students and focused on reproducing MoDL's results~\cite{3757-Aggarwal2019MoDL} utilizing its corresponding codebase~\cite{3757-modl} and dataset~\cite{3757-Aggarwal2019MoDLdataset}.  
  The model was trained with a variable density (VD) k-space sampling mask with an acceleration rate of R=6. The students additionally tested on self-generated VD mask following a Gaussian distribution and GRAPPA-style sampling masks, with results reported for R=4 and R=14.
  %B
  \item \emph{Team B:} Team B, comprising three students, tackled HUMUS-Net~\cite{3757-Fabian2022HUMUSNet} using its codebase~\cite{3757-humus_net} and single- and multi-coil data (fastMRI knee dataset~\cite{3757-zbontar2018fastmri}) with an acceleration rate of R=8. 
  Since this team had an additional student it was challenged to also evaluate the model on the brain multi-coil fastMRI dataset~\cite{3757-Aggarwal2019MoDLdataset}. 
  The students also compared the results to a Zero-filled approach.
  %C
  \item \emph{Team C:} The third team consisted of two students and focused on an "Untrained+Physics" model by Slavkova \emph{et~al.}~\cite{3757-Slavkova2023Untrained} utilizing its corresponding codebase~\cite{3757-cdr_mri} and dataset~\cite{3757-cdr_mri_dataset}.  
\end{itemize}
For all models, error maps were calculated as the difference between the reconstructed image and the fully sampled ground truth image. The evaluation metrics included structural similarity (SSIM) and peak signal-to-noise ratio (PSNR). 

All experiments were performed on a workstation featuring Ubuntu 24.04 LTS, an Intel Xeon W-2133 CPU, 30 GB RAM, and a single NVIDIA GeForce GTX 1080 GPU. 

The hackathon concluded with student teams presenting both the original papers and their reproduced results via posters within the seminar.%, the outcomes of which are detailed in the subsequent section.

\section{Results}
\label{3757-sec:results}
The results for the reproducibility hackathon for each team/paper are reported below.

\begin{figure}[!htbp]
	\centering
\begin{minipage}{\linewidth}
%\begin{table}
	\centering 
	\captionof{table}{Results for MoDL using in-domain and out-of-distribution (cross-domain) sampling masks.
	Image quality metrics for  Acc. rate: acceleration rate. SSIM: structural similarity. PSNR: peak signal-to-noise ratio.
	}
	\label{3757-tab:modl}
	\begin{tabular}{p{2.5cm} p{1.5cm} p{2.25cm} p{2.25cm} }
		\hline
		\textbf{Mask} & \textbf{Acc. rate} & \textbf{{SSIM}} & \textbf{{PSNR [dB]}}  \\
		\hline
		Original VD & R=6   & 0.92 $\pm$ 0.00 &  38.51 $\pm$  0.75  \\
		\hline
		Self-gen. VD & R=4 & 0.93 $\pm$ 0.00  & 40.06 $\pm$ 0.88  \\
		Self-gen. VD & R=14    & 0.85 $\pm$ 0.01  &  30.59 $\pm$ 1.01  \\ 
		\hline
		GRAPPA-style & R=4 & 0.87 $\pm$  0.01 & 30.56 $\pm$ 0.78   \\
		GRAPPA-style & R=14    & 0.52 $\pm$ 0.02  &  18.30 $\pm$  0.82 \\ 
		\hline
	\end{tabular}
%\end{table}
\end{minipage}
%
%\medspace

\vspace{5ex}

%\begin{figure}[t]
\begin{minipage}{\linewidth}
	\centering 
	\includegraphics[width=0.85\linewidth]{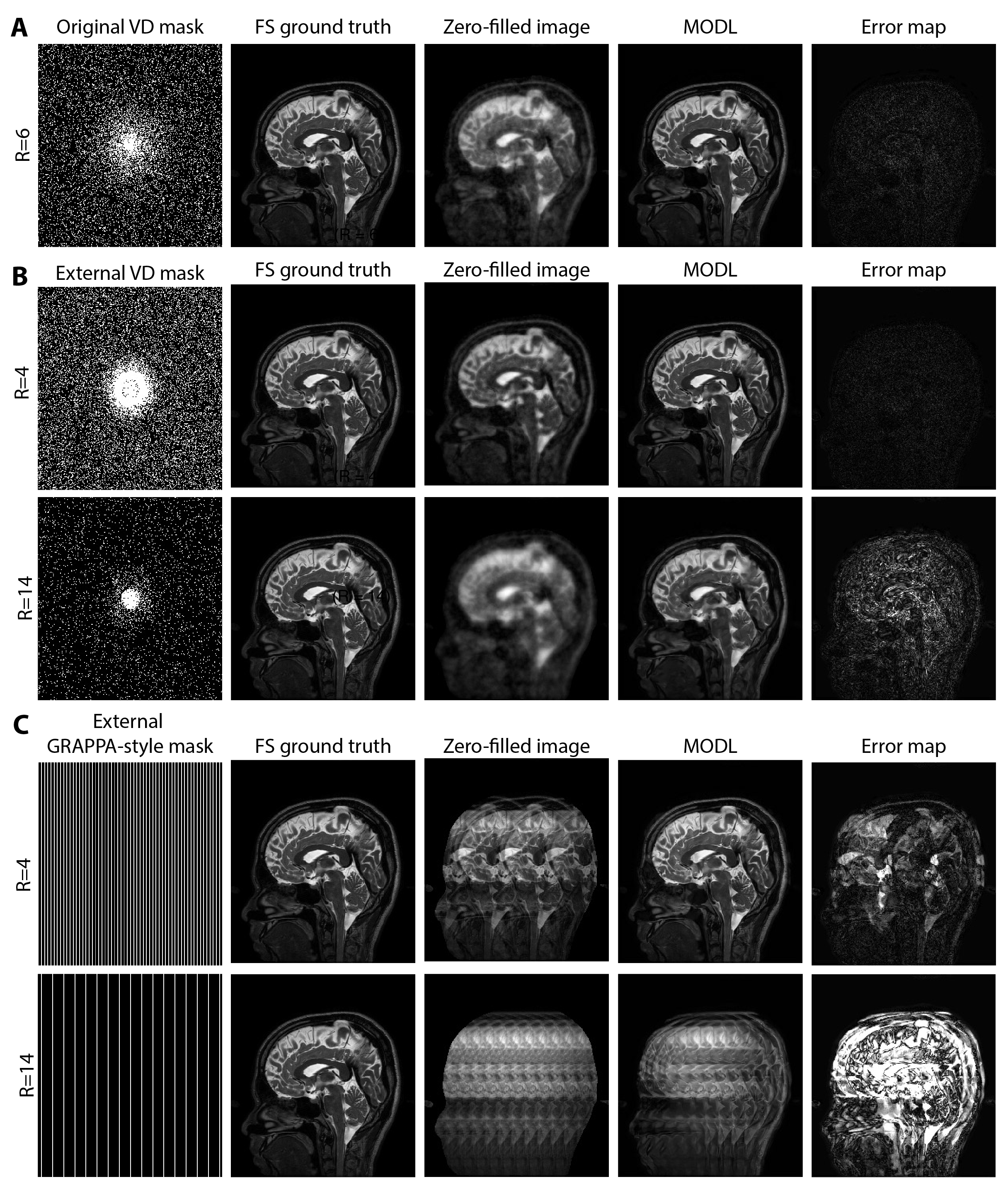}
	\caption{Results for MoDL using in-domain and out-of-distribution (cross-domain) sampling masks.
Fully-sampled (FS) ground truth, zero-filled image as well as MoDL reconstruction.
		(A) Same-domain results with an acceleration rate of R=6. 
		(B) Cross-domain results, tested in a self-generated variable density (VD) sampling mask, with acceleration rates of R=4 and R=14. 
		(C) Cross-domain results, tested in a self-generated GRAPPA-style sampling mask, with acceleration rates of R=4 and R=14.  
	}
	\label{3757-fig:modl}
%	\altText{Multi-panel figure comparing MRI reconstruction results under different undersampling masks and acceleration factors. 
%		Panel A shows an original variable-density (VD) sampling mask with acceleration R=6, alongside the fully sampled (FS) ground truth brain image, the corresponding zero-filled reconstruction, the MODL reconstruction, and an error map. 
%		Panel B shows external VD masks at accelerations R=4 and R=14, each with the FS ground truth, zero-filled image, MODL reconstruction, and error map. 
%		Panel C shows external GRAPPA-style Cartesian masks at R=4 and R=14 with the same set of reconstructions and error maps. 
%		Across panels, zero-filled images exhibit strong aliasing artifacts that increase with acceleration, while MODL reconstructions closely resemble the ground truth with reduced artifacts; error maps highlight residual reconstruction errors.
%	}
\end{minipage}
\end{figure}

%\begin{table}[H]
\begin{figure}[!htbp]
	\centering
	\begin{minipage}{\linewidth}
	\centering 
	\captionof{table}{Results for HUMUS-Net using in-domain (knee) and out-of-distribution, i.e., cross-domain (brain) target organs.  Image quality metrics. 
		PSNR: peak signal-to-noise ratio. 
		SSIM: structural similarity. }
	\label{3757-tab:humus}
	\begin{tabular}{p{1.5cm} p{4.0cm} p{1.5cm} p{1.5cm} }
		\hline
		\textbf{Dataset} & \textbf{Model} & \textbf{{SSIM}} & \textbf{{PSNR [dB]} } \\
		\hline
		Knee & HUMUS-Net (Train)        & 0.8870 & 36.28 \\
		Knee & HUMUS-Net-L (Train+val)  & 0.8912 & 36.71  \\
		\hline
		Brain & HUMUS-Net & 0.8871 & 31.9 \\
		Brain & Zero-filled     & 0.6832 & 25.9  \\
		\hline
	\end{tabular}
%\end{table}
\end{minipage}
%
%\medspace

\vspace{5ex}

%\begin{figure}[t]
\begin{minipage}{\linewidth}
%\begin{figure}[H]
	\centering
	\includegraphics[width=0.45\linewidth]{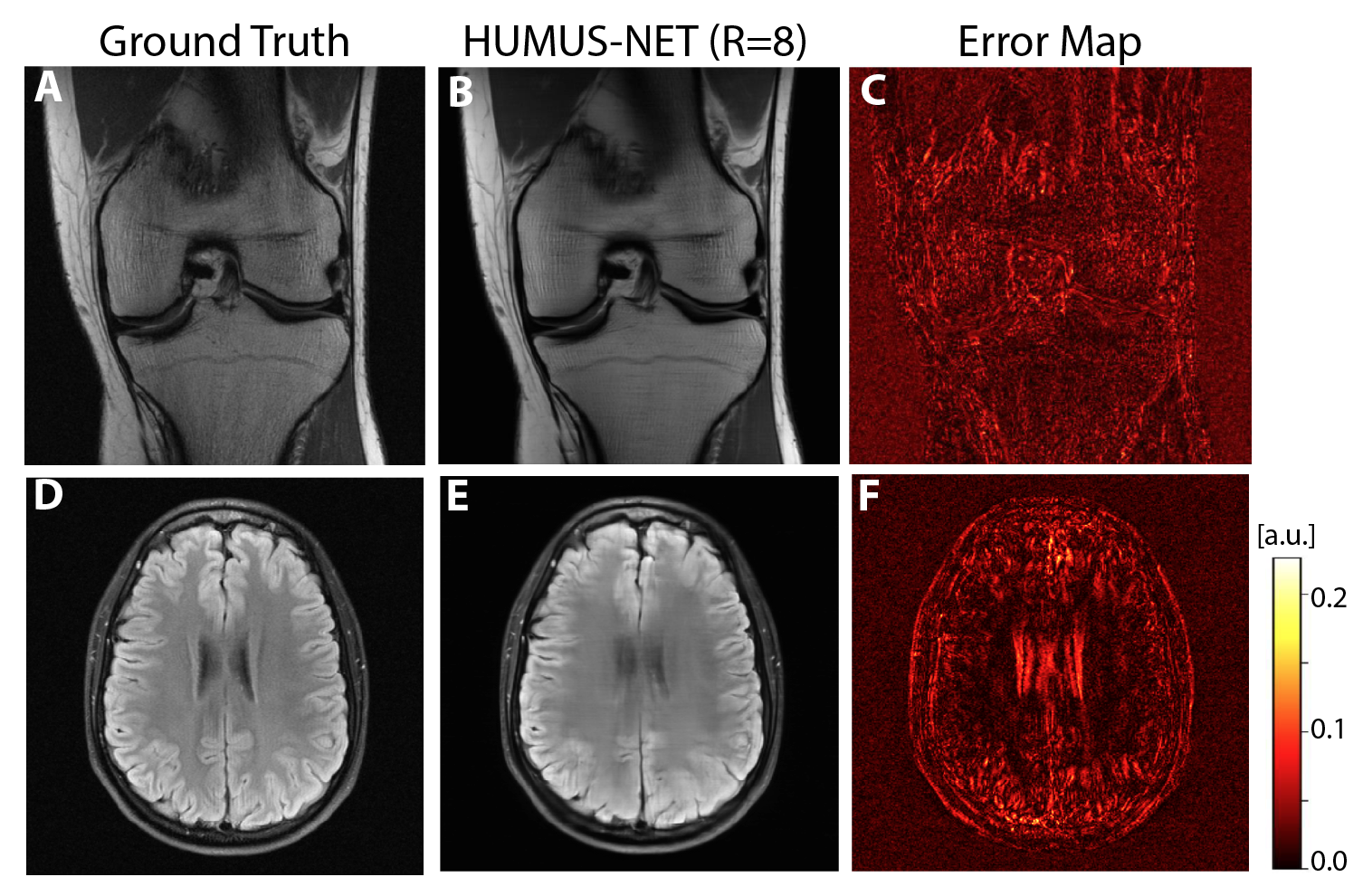}
	\caption{Results for HUMUS-Net using in-domain (knee) and out-of-distribution, i.e., cross-domain (brain) target organs. HUMUS-Net results together with the corresponding ground truth and error map. 
		(A-C): Same-domain reconstructions. % results.   
		(D-F): Cross-domain brain MRI reconstructions. %results.
	}
	\label{3757-fig:humus}
%	\altText{Six-panel figure comparing MRI ground truth images, HUMUS-Net reconstructions at acceleration R=8, and corresponding error maps. 
%		Panels A–C show a coronal knee MRI slice: A is the fully sampled ground truth, B is the HUMUS-Net reconstruction, and C is the error map displayed in red–yellow intensity. 
%		Panels D–F show an axial brain MRI slice: D is the fully sampled ground truth, E is the HUMUS-Net reconstruction, and F is the error map. 
%		A color bar indicates error magnitude in arbitrary units from 0.0 to approximately 0.2. 
%		The HUMUS-Net reconstructions closely match the ground truth in both anatomy types, with residual errors primarily along structural boundaries.
%	}
%}
\end{minipage}
\end{figure}

\paragraph{Team A: MoDL}
Team A could sucessfully reproduce the results from the MoDL paper.
Figure~\ref{3757-fig:modl}A shows representative images consisting of original VD mask, fully sampled ground truth, zero-filled image, MODL reconstructed image as well as the error map. Figure~\ref{3757-fig:modl}B and C demonstrate results from out-of-distribution cases, where the model was tested on a self-generated VD sampling mask, and a GRAPPA-style sampling mask. 
Table~\ref{3757-tab:modl} illustrates the image quality metrics involving PSNR and SSIM across in-domain acceleration rate of R=6, and out-of-distribution (i.e., cross-domain) results from self-generated VD and GRAPPA-style sampling masks at R=4 and R=14. 
The image quality metrics with self-generated VD masks were comparable to those from the model trained and tested with original VD mask across R=4 and R=14. 
When MoDL was tested on a self-generated GRAPPA-style mask, the PSNR and SSIM values decreased.

\paragraph{Team B: HUMUS-Net}
Other than suggested in the README files, the HUMUS-Net implementation did not support single-coil data.
However, the reproduction of the results for the multi-coil data was successful. 
The reconstruction results together with the corresponding fully-sampled ground truth images are shown in Figure~\ref{3757-fig:humus}. 
Table~\ref{3757-tab:humus} shows PSNR and SSIM results for in-domain (i.e., knee) and out-of-distribution (cross-domain) results. 
Interestingly, the SSIM values for the brain (out-of-distribution) data, was as good as for the in-domain trained knee data.
 
\paragraph{Team C: Untrained+Physics}
Unfortunately, the students were not able to reproduce the results.
While the paper presents promising results in untrained deep learning for dynamic MR image reconstruction, the code lacked several components necessary for robust and reproducible implementation. 

Documentation was insufficient: the README provided little guidance, and the requirements file omitted package versions, leading to execution failures due to deprecated methods. Critical elements, such as complex-valued loss handling, were missing, and the codebase contained numerous commented-out sections and non-functional placeholders that obscured the intended logic.

Reported expectations in the original work~\cite{3757-Slavkova2023Untrained} stated training times around 3–5 hours over 9,999 epochs with a gradually decreasing loss and convergence.
In our experiments, after ~11 hours of training, only ~5,050 epochs were completed and loss values remained high (e.g., loss ~~639,949 at epoch 5,049). 

The authors were contacted for clarification and indicated that the codebase is no longer actively maintained. The implementation relies on a previous version of PyTorch predating built-in support for complex numbers, which clarifies the handling of imaginary components as two separate channels. The authors recommended to create a Conda environment that uses
% the correct, 
 previous versions of the packages, and added that they were happy to help in that regard.  %Unfortunately, t
This was out-of-scope for this project.

\section{Discussion \& Conclusion}
\label{3757-sec:discussion}

In the reproducibility study, students were requested to mimic a realistic scenario of using pre-trained weights to evaluate the models and recreate one of the figures in the original study (i.e. no further training). 
%In case the students had extra time, they were asked to show additional out-of-distribution results.
When time permitted, students were asked to include additional out-of-distribution results. 
Out of three studies, two of them could be reproduced.
For MoDL and HUMUS-Net, results are in accordance with the published results. 
%Interestingly, i
For the out-of-distribution task it was found that the MoDL approach generalizes well to varying VD sampling patterns with different acceleration rates, which is in accordance with the authors theory behind weight sharing. 
The model performance suffered for self-generated GRAPPA-style sampling masks, possibly requiring training from scratch. 
HUMUS-Net yielded good generalization among out-of-distribution target organs (e.g., brain). 
%The students were not able to reproduce the reconstructions for the Untrained+Physics model.
The Untrained+Physics model could not be reproduced.

There were a few reasons as to why the teams could complete the task and do further testing with the first two studies. 
The studies that could be reproduced had:
(i) proper  documentation with clear guidance on their README files;
(ii) complete dependency list of required libraries and Python versions.
The third study could not be reproduced due to an outdated codebase and README, insufficient information on training configurations, inadequately commented code, and missing training checkpoints.
This highlights the importance of the FAIR and FUTURE-AI principles and commitment to reproducibility.
%

%The main limitation of the hackathon was the strict four-hour time constraint, as it was conducted only as a subpart of a seminar.
%In the future, we could do a full seminar course with more students on reproducibility, including model training, in which the students would be permitted to have longer durations for implementation and experiments. A full study on reproducibilty is more challenging to set up, yet more interesting. 
The main limitation of the hackathon was the strict four-hour time constraint, as it was conducted as part of a seminar. Future work could involve a full seminar course on reproducibility with more students, including model training and extended time for implementation and experiments. While more challenging to organize, a full reproducibility study would be more informative.

Overall, this work demonstrates a reproducibility hackathon on MR image reconstruction in a teaching environment. 
We conclude that clean, well-documented, and reproducible code should be a prerequisite for publication to ensure transparency and to fight the reproducibility crisis~\cite{3757-Stupple2019Reproducibility}.

%\begin{acknowledgement}
%	This work is supported by the Konrad Zuse School of Excellence in Reliable AI (relAI).
%	Dr. Sevgi Gokce Kafali is sponsored by Alexander von Humboldt Foundation in Germany.
%\end{acknowledgement}

%\printbibliography
% Specification of your bibliography file (with file extension)
%\addbibresource{bvm_hackathon.bib}

\bibliographystyle{plain} % We choose the "plain" reference style
\bibliography{bvm_hackathon} % Entries are in the refs.bib file

\end{document}